\documentclass[12pt,a4paper]{article}

\usepackage[T1]{fontenc}
\usepackage[utf8]{inputenc}
\usepackage{times}
\usepackage[margin=1in]{geometry}
\usepackage{amsmath,amssymb}
\usepackage{graphicx}
\usepackage{float}
\usepackage{booktabs}
\usepackage{array}
\usepackage{xcolor}
\usepackage{listings}
\usepackage{microtype}
\usepackage{placeins}
\usepackage[numbers,sort&compress]{natbib}
\usepackage{url}
\usepackage[hidelinks]{hyperref}

\newcommand{\figorbox}[1]{%
  \IfFileExists{#1}{\includegraphics[width=\linewidth]{#1}}{%
    \fbox{\parbox[c][3.2cm][c]{0.96\linewidth}{\centering\ttfamily\detokenize{#1}}}}%
}
\definecolor{promptbg}{gray}{0.95}
\lstdefinestyle{prompt}{basicstyle=\ttfamily\small,breaklines=true,frame=single,
  framesep=4pt,rulesep=0pt,backgroundcolor=\color{promptbg},columns=fullflexible,
  keepspaces=true,xleftmargin=0pt,xrightmargin=0pt}
\newenvironment{promptlabel}[1]
  {\smallskip\noindent\textbf{#1}\par\nobreak\smallskip
   \noindent\hspace*{5pt}\begin{minipage}{\dimexpr\linewidth-10pt\relax}}
  {\end{minipage}\smallskip}
\title{\vspace{-1.5em}Large language models represent suicidality\\
before they causally use it}

\author{
Nafiz Ahmed$^{1}$,
Sarah Sharif$^{1}$,
Dingjing Shi$^{2}$,
Mike Banad$^{1,\ast}$\\[6pt]
{\small $^{1}$Intelligent Neuromorphic and Quantum Understanding for Innovative Research and}\\
{\small Engineering (INQUIRE) Lab, School of Electrical and Computer Engineering,}\\
{\small University of Oklahoma, Norman, OK 73019, USA}\\
{\small $^{2}$School of Psychology, Georgia Institute of Technology, Atlanta, GA 30332, USA}\\[4pt]
{\small $^{\ast}$Corresponding author: bana@ou.edu}\\[4pt]
{\footnotesize ORCID:\
N.A.~\href{https://orcid.org/0009-0004-6100-1995}{0009-0004-6100-1995};\
S.S.~\href{https://orcid.org/0000-0002-4667-6650}{0000-0002-4667-6650};}\\
{\footnotesize
D.S.~\href{https://orcid.org/0000-0002-5652-3818}{0000-0002-5652-3818};\
M.B.~\href{https://orcid.org/0000-0001-7339-810X}{0000-0001-7339-810X}}
}
\date{}

\begin{document}
\maketitle

\begin{abstract}
\noindent Large language models are proposed to flag suicidal content, but representing a
concept and acting on it are distinct. A suicidality feature is decodable from 0.5B
parameters upward, while the behavior that uses it emerges only in the low billions. Where
it emerges, it rests on a compact mid-network feature: in Llama-3.1-8B-Instruct this
direction decodes suicidal from non-suicidal text above a matched lexical baseline and
survives removal of the 38 most label-predictive keywords. Projecting it out drops the
model's ranking from AUC $0.975$ to $0.832$, while an equal-norm random direction does not;
ablating a low-dimensional subspace degrades it further, though no stable rank is
recoverable. The direction transfers to
two further corpora, replicates in Qwen and Mistral, and tracks clinician-assigned risk.
Steering raises an off-target readout as much as the target across five concepts, so
sufficiency claims from yes/no readouts are confounded by generic salience.
\end{abstract}

\noindent High-stakes decisions are increasingly delegated to LLMs, and interpretability is
invoked to justify trusting them. Most interpretations, however, are post hoc: they report
which inputs accompany a decision, not which internal computation the decision causally
depends on. When the stakes are high, an explanation that cannot be falsified is not enough:
we need causal evidence that a behavior rests on the feature an interpretation names, and
a discipline for separating trustworthy causal claims from plausible but untested ones. We
develop such a discipline and instantiate it on a safety-critical behavior: detecting
suicidal content.

LLMs are increasingly proposed for mental-health triage, including the detection of suicidal
content~\citep{rezaei2026mentalhealth}. Black-box classifiers already detect suicidal
ideation on Reddit at high accuracy, whether fine-tuned
transformers~\citep{li2024suicidereddit} or domain-pretrained
encoders~\citep{ji2022mentalbert}, but they report what is predicted, not why. The stakes are concrete: a series of wrongful-death
suits now allege that deployed conversational systems reinforced rather than recognized
users' suicidal ideation, several reaching confidential settlements in early
2026.\footnote{E.g.\ \emph{Garcia v.\ Character Technologies} (M.D.\ Fla., 2024) and
\emph{Raine v.\ OpenAI} (Cal.\ Super.\ Ct., 2025); related cases settled January 2026.}
Before such systems can be trusted, we should understand what a model uses to make these
judgments: a specific, interpretable representation of suicidality, or a shallow proxy such
as keyword presence. Mechanistic interpretability~\citep{zhao2024explainability} offers
tools to answer this: linear probes locate what is decodable from intermediate
activations~\citep{alain2017probes,belinkov2019analysis}, while ablation, steering, and
patching test whether it is causally used, decodability alone establishing no such
use~\citep{belinkov2022probing}. But applying them to a named concept invites a recurring failure:
attributing a ``mechanism'' to a behavior the model never reliably performs, or treating a
steering effect as proof a feature is sufficient. We turn a single discipline, \emph{no
mechanistic claim about a behavior the model does not demonstrably perform}, into a
falsifiable, reusable workflow, and use suicidality detection as the case study that
stress-tests it.

This paper makes three contributions. (1)~A causal account of how an LLM detects suicidal
content: one mid-network feature, semantic rather than lexical, causally necessary, and
low-dimensional, each established against a matched control, recurring across three model
families and three corpora and tracking clinician-assigned risk. To our knowledge this is the
first causal, cross-model account of an LLM's suicidality-detection behavior; the
contribution is the mechanism, not a new classifier. (2)~A scale dissociation between
encoding and use: the representation is present from 0.5B parameters, but the model only acts
on it in the low billions, so a probe that lights up in a deployed model certifies nothing
about what that model does. This is what makes necessity testing, rather than decoding, the
operative check for an audit, and it has a within-model consequence: at 8B the internal
readout flags cases the model's own answer misses. (3)~A steering-sufficiency confound and the control that
exposes it: across five concepts, adding a difference-of-means ``concept direction'' read
through a yes/no logit drives \emph{every} readout, including an off-target one, to nearly
the same level, while a magnitude-matched random direction does not; steering-based
sufficiency claims via polar readouts are thus confounded by a generic salience push that only
an off-target control exposes. This is the result that generalizes furthest beyond
suicidality.

Holding these claims to that standard requires a discipline, which we call \emph{validation
gating} and apply throughout: \emph{no mechanistic claim about a behavior the model does not demonstrably
perform}. Concretely, a behavioral gate admits a concept only after the model ranks the
classes above a lexical baseline, since establishing that a model reliably performs a task is
a prerequisite to trusting any reading of
it~\citep{ziems2024css,chang2024evaluation}; every rung of the analysis that follows is tested against a
matched control, namely decodability, necessity (directional ablation), intrinsic dimension
(rank-$k$ ablation) and sufficiency (steering); and generality is checked by decoding and
causal transfer across datasets and models. Because every rung is falsifiable, the procedure
returns negatives as well as confirmations, and here it returns four: an implicit-intent task
that fails the gate outright, steering sufficiency, the rank of the necessary subspace, and a
sparse head- or edge-level circuit.
Figure~\ref{fig:framework} lays out the study end to end; Table~\ref{tab:summary} collects every
property against its control. Extended methods, robustness checks, and the localization
analysis are in the Supplementary Information (SI).

\begin{figure}[p]\centering
  \figorbox{figures/research_overview.pdf}
  \caption{\textbf{The study in one view.} \textbf{a}, Four corpora, one prompt template and
  three model families, so that a result has to survive a change of corpus and a change of
  model. \textbf{b}, The portion of the network the causal claims rest on. Of 32 layers the
  probe peaks at 15 and every causal read and write is at layer 16; inside that one layer,
  a low-dimensional subspace of 4096. \textbf{c}, The four interventions applied there. Light red marks
  every change made to the model, and marks nothing else. \textbf{d}, Each rung against its
  matched control on one behavioral-AUC scale: the heavy tick is the measurement, the light
  tick its control, and the bar between them is the evidence. Filled where the separation
  holds, outlined where it does not. The three outlined rungs are three of the four
  negatives; the fourth, the rank of the necessary subspace, is a property of the
  low-dimensional rung rather than a rung of its own. Values in Table~\ref{tab:summary}; the readout and the geometry of the
  interventions are drawn in Figs.~\ref{fig:readout} and~\ref{fig:geometry}.}
  \label{fig:framework}
\end{figure}

\section*{Results}

Throughout, a claim is admitted only after the model demonstrably performs the
behavior: the classes must be ranked above a matched lexical baseline before any internal
analysis begins. Binary suicide detection on Suicide-Watch-232K~\citep{komati2021suicidewatch}
clears that gate, and is the behavior analyzed below. Implicit suicidal intent does not clear
it, and is not analyzed; that result, and a second null at the head and edge level, are
reported together under \emph{What the analysis rules out}.

\subsection*{A single mid-network feature carries the judgment}
\begin{figure}[tbp]\centering
  \includegraphics[width=0.745\linewidth]{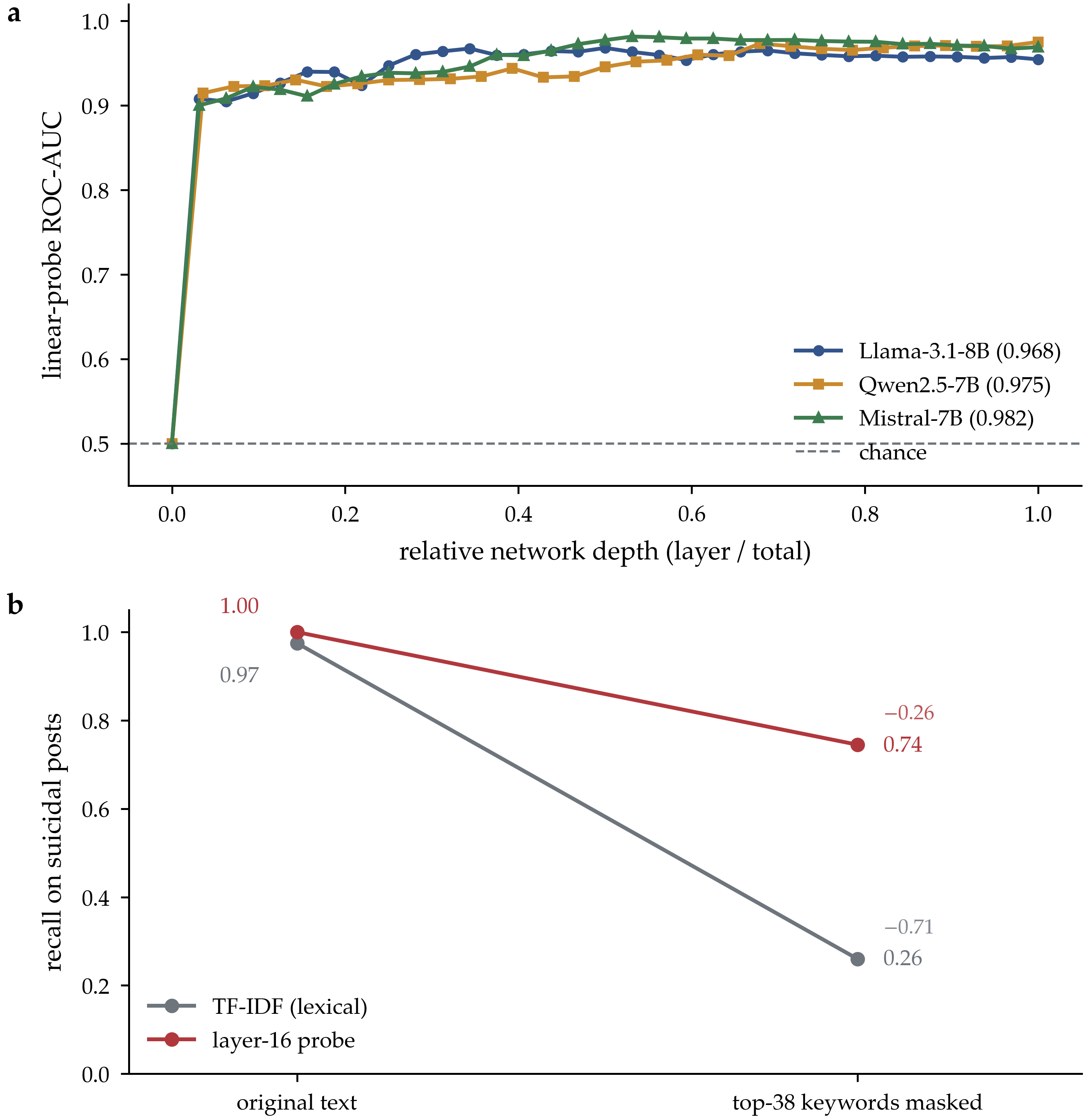}
  \caption{\textbf{The feature emerges mid-network, and is semantic rather than lexical.}
  \textbf{a}, Linear-probe AUC for suicide vs.\ non-suicide against layer depth, in all
  three model families. Decodability rises through the early layers, peaks in the
  mid-network band (layers 12--16 in Llama), and stays high thereafter. \textbf{b}, Recall
  on suicidal posts before and after masking the 38 most label-predictive TF-IDF tokens.
  The lexical TF-IDF model collapses ($0.97\to0.26$); the layer-16 probe barely moves
  ($1.00\to0.74$).}
  \label{fig:emergence}
\end{figure}

On Suicide-Watch-232K~\citep{komati2021suicidewatch}, the model ranks suicidal above
non-suicidal text well above its lexical baseline (behavioral AUC $0.961$ vs.\ matched
TF-IDF $0.894$), despite an affirmative response bias that makes argmax accuracy
uninformative (SI Section~\ref{sec:limitations}). A linear
probe reaches AUC $0.969$ at layer 15, again above the lexical baseline. The feature is
semantic rather than lexical: removing the 38 most label-predictive keywords collapses
TF-IDF recall on suicidal posts from $0.975$ to $0.285$, but probe recall falls only to
$0.740$ (Fig.~\ref{fig:emergence}b), roughly a threefold smaller drop. Because the 232K negative class is drawn from a different subreddit register, we treat
these AUCs as an upper bound and rely instead on a register-matched contrast (suicide vs.\
depression, both distressed mental-health text, $n=150$): the feature still clears its
lexical baseline (probe AUC $0.796$, $p=0.04$) and remains causally necessary there
(ablation $0.681\to0.466$ against a random-direction control at $0.686$; SI
Table~\ref{tab:summary-full}),
so it reads as a suicidality signal rather than a generic-distress or register proxy, though
the margin is modest at this sample size.

\subsection*{The feature is necessary, but its rank is not identified}
Projecting the suicidality direction out of the residual stream drops the model's ranking
from AUC $0.975$ to $0.832$, while an equal-norm random direction leaves it untouched
($0.975$, a drop of $0.000$); a disjoint downstream detector shows the same asymmetry
(F1 $0.918\to0.690$ for the suicidality direction, unchanged at $0.918$ for the random one;
SI Table~\ref{tab:downstream}). The single direction, however, is not the whole story:
ablating a low-dimensional subspace per layer degrades behavior further, and on the
downstream detector subspace ablation drives F1 to $0.000$, silencing it completely.

How many directions the behavior needs is a question our design cannot answer, and we report
that as a negative. A dense sweep in fp16 over $k=1$--$10,12,16$ at $n=200$/class with three
seeds and bootstrap intervals degrades behavior at every rank tested, from a baseline of
$0.963$ to between $0.53$ and $0.77$ depending on $k$, against a random rank-$k$ subspace
that is inert everywhere ($0.950$--$0.967$ across all ranks and seeds; Fig.~\ref{fig:necessity}).
But the curve neither descends to chance nor stays there. It is non-monotone, seed-dependent
(the spread across seeds reaches $0.38$ AUC at $k=16$ and $0.32$ at $k=3$), and at some ranks
falls \emph{below} chance ($0.331$ at $k=16$ in one seed), which indicates the ablation
rotates the decision axis rather than removing the information. Estimating $k^\ast$ as the
smallest rank not significantly above chance therefore returns $3$, $3$ and $10$ across the
three seeds, and the estimator cannot distinguish an inverted readout from an erased one.
We conclude that necessity is a property of a low-dimensional subspace rather than of any
single direction, and that the dimension of that subspace is not identifiable by rank-$k$
ablation scored with a sign-blind AUC threshold. The sweep, the pre-registered decision rule
it was scored against, and the per-seed intervals are in SI Section~\ref{sec:ranksweep}. The same
gate-necessity-rank ladder run on a second concept, toxicity, reproduces the pattern
(necessity $0.910\to0.567$, random $0.913$; SI Section~\ref{sec:secondconcept}), showing the protocol
transfers, not only the suicidality result.

\begin{figure}[tbp]\centering
  \figorbox{figures/fig2_rank_sweep.png}
  \caption{\textbf{The rank of the necessary subspace is not identified.} Dense rank-$k$
  ablation in fp16 at $n=200$/class, one line per seed, with 2{,}000-sample bootstrap 95\%
  intervals. Every rank degrades behavior well below the unablated baseline (blue) while the
  random rank-$k$ control (grey) is inert throughout, so the subspace is doing causal work.
  But the curve is non-monotone and strongly seed-dependent, and at several ranks it falls
  below chance, which marks an inverted ranking rather than an erased one. No stable $k^\ast$
  is recoverable: per-seed estimates are $3$, $3$ and $10$. Single-direction necessity is
  reported in the text ($0.975\to0.832$, random $0.975$).}
  \label{fig:necessity}
\end{figure}

\subsection*{Steering raises every readout, not the targeted one}
Ablation establishes that the feature is necessary; the converse rung asks whether adding it
is sufficient, and this is where the analysis returns its first negative.
Adding the suicidality direction to non-suicidal text (coefficient $c=1$) pushes $100\%$ of
posts past the class-separating threshold, against $8\%$ at baseline and $14\%$ for an
equal-magnitude random direction, the standard evidence offered for steering
``sufficiency.'' A stronger, off-target control overturns that reading: on the same steered
pass, an unrelated question (``is this about food?'') is driven \emph{past} the on-target
question ($16.5$ vs.\ $9.6$ logits of shift), despite starting well below it. This confound
generalizes. Across five concepts (suicidality, toxicity, sentiment, and two neutral news
topics), steering any one of them lifts \emph{every} yes/no readout by a comparable amount
regardless of the question asked, and for four of the five the largest shift falls on the
off-target food question, while a magnitude-matched random direction moves every readout by
less than $1.5$ logits (Fig.~\ref{fig:confound}). No direction is concept-selective
(0/5). Steering-based sufficiency claims via a polar (yes/no) readout are therefore
confounded by injection magnitude, not concept match, unless an off-target control is
included, a lesson for interpretability work well beyond this case study. Steering vectors
are already known to be unreliable and input-dependent in effect
size~\citep{tan2024steering,rimsky2024caa}; what the off-target control adds is a
selectivity test specific to polar readouts, where a magnitude push alone can pass for a
concept match.

\begin{figure}[tbp]\centering
  \figorbox{figures/steering_salience.png}
  \caption{\textbf{The steering-salience confound is general.} Left: Yes/No readout level on
  neutral substrate at $c=1$ (rows = steering direction, columns = readout question); every
  concept row is elevated uniformly, while the random-direction row stays low and varied.
  Right: within-row spread, near zero for every concept direction against the $\sim$9-logit
  spread a random direction preserves.}
  \label{fig:confound}
\end{figure}

\subsection*{The feature transfers across corpora and model families}
The direction fit on 232K transfers without retraining, both by decoding (SDCNL AUC
$0.963$; SWMH AUC $0.917$) and causally (a random direction does nothing on
either; Fig.~\ref{fig:general}). It also
replicates across model families, and more sharply than in Llama: ablating the direction
drops behavioral AUC $0.96\to0.69$ in Qwen2.5-7B-Instruct and $0.96\to0.59$ in
Mistral-7B-Instruct-v0.3, against equal-norm random directions that leave each unchanged
(Fig.~\ref{fig:scale}; SI Table~\ref{tab:crossmodel}). Expressed as the share of
above-chance signal removed, single-direction necessity is weakest in the primary model
($0.30$) and strongest in Mistral ($0.80$), which is why the subspace result above, not the
single direction, carries the mechanistic claim.

\begin{figure}[tbp]\centering
  \figorbox{figures/fig5_generalization.png}
  \caption{\textbf{The feature transfers across corpora.} The direction fit on
  Suicide-Watch-232K is applied without retraining to SDCNL and SWMH. Decoding transfer:
  it separates both held-out corpora (AUC $0.962$ and $0.919$). Causal transfer: ablating
  that same 232K direction collapses behavior on each ($0.947\to0.632$ on SDCNL,
  $0.932\to0.631$ on SWMH), while an equal-norm random direction does nothing.}
  \label{fig:general}
\end{figure}

\subsection*{The internal axis tracks clinician-assigned risk}
Projecting text onto the suicidality direction yields a monotonic ordering (neutral $<$
depression $<$ implicit ideation $<$ explicit suicide). On the expert-annotated UMD
dataset~\citep{shing2018umd,zirikly2019clpsych}, the same 232K direction recovers the
clinician ordering monotonically (Spearman $\rho=0.56$, $p=8\times10^{-22}$), statistically
indistinguishable from the model's own behavioral readout ($\rho=0.56$; TOST $p=0.01$; Fig.~\ref{fig:umd}). UMD annotates the \emph{user}, not the post, so the unit of analysis is one user
($n=245$), with that user's SuicideWatch posts concatenated (Methods). Because higher-risk
users also write more (aggregated length vs.\ severity $\rho=0.25$, $p=1\times10^{-4}$), we
partial length out: the association is unchanged ($\rho=0.57$ given characters, $0.55$ given
post count), so the axis tracks risk rather than verbosity. Suicide risk has been cast as an
ordinal severity scale in its own right~\citep{sawhney2021ordinal}, and that is the structure
the axis recovers internally, consistent with concepts being directions in activation space
whose ordinal attributes are encoded
monotonically~\citep{park2024linear,heinzerling2024monotonic}. The internal axis therefore
offers interpretable, ablatable access to a graded risk judgment.

Concurrent work reaches the decoding half of this result independently and by a different
route: \citet{zhu2026symptomvectors} build difference-of-means symptom vectors in
Gemma-3-27B from clinical instruments and show that projections onto a suicidality axis
preserve clinician-annotated rank ordering on held-out narratives, and
\citet{ghanadian2026beyondaccuracy} recover topic-selective features from a sparse
autoencoder over a fine-tuned classifier on this same UMD corpus. Neither intervenes: both
are projection and separability analyses, and both note the absence of causal tests as a
limitation. That is the gap the necessity results above are meant to close, and it is why we
treat decodability as the entry condition to the analysis rather than as its conclusion.

\begin{figure}[tbp]\centering
  \figorbox{figures/umd_severity.png}
  \caption{\textbf{The internal axis tracks clinician-assigned risk.} Left: mean
  projection of $n=245$ UMD users onto the 232K suicidality direction, by
  clinician-assigned risk level (a$<$b$<$c$<$d), with bootstrap 95\% CIs; the ordering is
  monotonic (Spearman $\rho=0.56$, $p=8\times10^{-22}$). Right: separation of high-risk
  (c,\,d) from low-risk (a,\,b) users by the feature projection and by the model's own
  behavioral readout, with bootstrap 95\% CIs and the chance line marked; the direction
  transfers at AUC $0.81$ with no retraining, statistically indistinguishable from the
  readout at $0.82$.}
  \label{fig:umd}
\end{figure}

\subsection*{Models encode suicidality before they use it}
Across scale, the representation is present everywhere (probe AUC
$0.94$--$0.98$ from 0.5B to 8B in both families), but the behavioral readout emerges only
with scale: at $\approx$1B, models encode suicidality strongly yet classify at chance; from
$\approx$1.5B upward the behavioral readout returns, while clean single-direction necessity
is only reliable from 7--8B. A positive control rules out the trivial explanation (sub-2B
models drive other yes/no readouts, such as sentiment, at AUC $0.93$--$0.96$, but score at
chance on suicidality despite a strong probe): the small-model gap is genuine non-use, not
formatting incompetence. Encoding a concept and computing with it are therefore distinct and
separately measurable.

\begin{figure}[tbp]\centering
  \figorbox{figures/fig8_crossmodel_bars.png}
  \caption{\textbf{Encoding versus use across scale and family.} Probe AUC (encoding) stays
  high from 0.5B to 8B in Llama, Qwen, and Mistral, while behavioral AUC (use) and
  single-direction necessity emerge only at larger scale. Per-model numbers in SI
  Table~\ref{tab:crossmodel}.}
  \label{fig:scale}
\end{figure}

\subsection*{The internal readout catches cases the model's own answer misses}
The dissociation between encoding and use has a consequence at 8B, not only at 1B. On
$1{,}000$ held-out posts per class, with one decision threshold per readout calibrated on a
disjoint split, the model's behavioral readout reaches AUC $0.961$ and recall $0.796$,
missing $204$ of the $1{,}000$ suicidal posts. A linear probe on the same forward passes
reaches AUC $0.973$ and recall $0.930$, missing $70$. Because a readout that simply flags
more posts would recover misses for free, we hold the operating point fixed rather than
comparing each readout at its own calibrated threshold. Constrained to flag exactly as many
posts as the model does ($839$ of $2{,}000$), the probe attains both higher recall ($0.807$
vs.\ $0.796$) and higher precision ($0.962$ vs.\ $0.949$), and $45.6\%$ of the model's $204$
missed suicidal posts fall inside its alarm set: it flags a differently chosen set of the
same size, not a larger one. Constrained instead to the model's precision ($0.949$), it
reaches recall $0.872$ and recovers $57.4\%$ of the misses. The effect transfers without
refitting to SDCNL and SWMH. The internal state is therefore not merely a redescription of
the output.

Three limits belong with that claim. The advantage is carried by the trained probe, not by
the difference-of-means direction that supports the causal results above: at a matched alarm
rate the raw projection recovers only $14\%$ of the misses, so the readout that is most
useful for monitoring is not the one that is most interpretable. Within the set of cases the
model gets wrong the probe cannot rank them (AUC $0.52$), so the gain is a better operating
point rather than better discrimination on the hard cases. And this is a monitoring result
on static posts under a fixed prompt, not a validated clinical instrument.

\subsection*{What the analysis rules out}
Besides steering sufficiency above, two claims a less disciplined analysis would have asserted
do not survive: one is foreclosed by the behavioral gate before any internal analysis begins,
the other fails its control.

\noindent\textbf{Implicit intent is not represented.} On DeepSuiMind~\citep{li2025implicit}, Llama-3.1-8B-Instruct cannot separate implicit
suicidal intent (Death-Me) from matched distress (Life-Not-Me): behavioral AUC $0.557$ and a
32-layer linear probe reaching at most AUC $0.570$, both below the $0.599$ lexical baseline,
while a positive control (distress vs.\ neutral text) reaches AUC $1.000$. The model encodes
distress, not implicit intent, so the framework forbids any mechanistic claim about this
task and we do not analyze it further (SI Section~\ref{sec:mechfigs}).

\noindent\textbf{No sparse head- or edge-level circuit.} Per-head knockout over all 1,024
heads ($n=60$/class), attribution patching~\citep{syed2023attribution,kramar2024atp}, and exact single-edge path-patching over all
2,145 causal edges agree: no single head or small edge set carries the behavior. The largest
single-head effect is an AUC drop of $0.013$ with a bootstrap CI spanning zero
($[-0.019,0.051]$), no head survives FDR correction ($0/1024$ at $q<0.05$), and the mean
effect across all heads is $+0.0001$. The identity of the nominal top head is itself
unstable across sample size (L11H1 at $n=20$, L9H17 at $n=60$), as expected when there is no
head to find. Edge attribution tells the same story: of the $2{,}145$ edges, $249$ are needed
to account for half the attribution mass, the twenty strongest carry $9.2\%$ of it (against a
$0.9\%$ uniform null), and the twelve strongest recover only $0.27$ of the clean--corrupt
gap under exact patching. Attribution is thus an order of magnitude above uniform yet far
from sparse. A sparse autoencoder~\citep{bricken2023monosemanticity,cunningham2023sparse}
at the same layer instead recovers a few sharply suicide-selective atoms, the monosemantic
counterpart of the low-dimensional subspace above, as expected if features are stored in
superposition~\citep{elhage2022superposition}. Our causal claims are therefore made at the
direction/subspace level, not individual heads or edges, and no sparse feature circuit of
the kind recoverable for narrower behaviors~\citep{marks2024sparse} appears here (full localization figures in
SI Section~\ref{sec:mechfigs}).

\begin{table}[t]\centering
\setlength{\tabcolsep}{4pt}\footnotesize\renewcommand{\arraystretch}{0.95}
\begin{tabular}{p{0.19\linewidth} p{0.30\linewidth} p{0.22\linewidth} p{0.19\linewidth}}
\toprule
Property & Result & Control / baseline & Outcome \\
\midrule
Intent not represented & behavioral / probe AUC $0.557$ / $0.555$--$0.570$ & TF-IDF $0.599$ & gated out \\
Decodable, semantic & probe AUC $0.969$; recall $1.00\to0.74$ under masking & TF-IDF $0.894$; TF-IDF recall $0.98\to0.29$ & confirmed \\
Necessary & ablation AUC $0.975\to0.832$; downstream F1 $0.92\to0.69$ & random dir.\ $0.975$ (drop $0.000$) & confirmed \\
Low-dimensional & every rank degrades ($0.963\to0.53$--$0.77$); downstream F1 $\to0.00$ & random rank-$k$ $0.950$--$0.967$ & confirmed \\
Rank identified & $k^\ast=3,3,10$ across seeds; curve non-monotone, dips below chance & sign-blind AUC threshold & \textbf{not identified} \\
Sufficiency & steering flip rate $100\%$; off-target shift exceeds on-target & random $<1.5$ logits & \textbf{not established} \\
Generalizes & transfer AUC $0.963$/$0.917$; Qwen $0.960\to0.328$ & random dir.\ inert & confirmed \\
Scale dissociation & probe $0.94$--$0.98$ at $0.5$--$8$B; behavior from $\approx1.5$B & chance below $\approx1$B & confirmed \\
Internal monitor & at a matched alarm rate, probe recovers $46\%$ of the model's false negatives & same $839$ alarms; precision $0.962$ vs.\ $0.949$ & confirmed \\
Severity gradient & $\rho=0.56$ vs.\ clinician risk ($n{=}245$ users) & $\rho=0.57$ given length & confirmed \\
Sparse circuit & top head drop $0.013$, CI $[-0.019,0.051]$; $0/1024$ after FDR & distributed null & \textbf{not found} \\
\bottomrule
\end{tabular}
\caption{\textbf{Evidence at a glance.} Every row carries a matched control. SI
Table~\ref{tab:summary-full} gives the full 18-row scorecard; headline AUCs carry
2,000-sample bootstrap 95\% CIs, reported with each result in the text.}
\label{tab:summary}
\end{table}

\section*{Discussion}
The model's suicide-detection behavior rests on a single, compact, mid-network feature that is
semantic rather than lexical, causally necessary, low-dimensional, and stable across corpora
and model families.
Encoding is not the same as use: the representation is strong at every scale we tested, yet
the model only acts on it from roughly 1.5B parameters upward, which is why, for auditing a
deployed model, necessity tests, not the probe, carry the actionability claim. Each
ingredient of this account exists in prior
work~\citep{li2025implicit,basu2026actionability,arditi2024refusal,heinzerling2024monotonic,beaglehole2025universal};
our contribution is to assemble them into one causal, cross-model account of a
safety-relevant concept. Interpretability applied to clinical language models has so far
reported feature attribution rather than causal mechanism~\citep{rezaei2025shap,rezaei2026pharma,ahmadi2024transparent};
sparse-autoencoder searches surface candidate mental-health features in open
models~\citep{grabb2024psaechiatry} but discover and steer features rather than validate a
model-performed behavior through a gate; and the nearest mechanistic work on affect-laden
internal states targets pain--pleasure decisions rather than clinical
risk~\citep{bianco2026painpleasure}. What we add is a causal account of a
safety-relevant concept behind a behavior-baseline gate, falsifiable at every rung: here it
falsified four plausible claims a less disciplined analysis would have asserted
(implicit-intent representation, steering sufficiency, a specific subspace dimension, and a
sparse head/edge circuit).

Two implications generalize beyond suicidality. For auditing, a low-rank, faithful feature
is a practical handle: it can be monitored or ablated to probe a deployed model's reliance
on a safety-relevant cue without retraining, and cross-model consistency suggests such
handles transfer. But a probe that lights up does not certify reliance, so the handle must
rest on a causal audit, not decodability alone. For methodology, researchers evaluating
concept directions and steering more broadly should adopt off-target and behavioral controls
before drawing causal or safety conclusions; the steering-salience confound we document was
present in every one of five unrelated concepts we tested, suggesting it is a property of
the intervention, not of suicidality specifically.

The evidence has clear scope. All mechanistic numbers are measured in full precision
(fp16); an earlier 4-bit run reproduces every qualitative conclusion but overstates
single-direction necessity, dropping AUC to $0.66$ where fp16 gives $0.83$, which is part of
why we locate the claim at the subspace rather than the direction (SI Table~\ref{tab:precision}). The binary task's negative class carries
a register shift, so we treat the register-matched contrast, not the headline AUC, as the
specificity evidence, and it is modest ($p=0.04$, $n=150$). On expert-annotated data, the
internal severity axis matches the model's own behavioral readout on the full sample, and
exceeds it only on the subset the model gets wrong. Steering establishes only a
magnitude-controlled effect, not concept-selective sufficiency. The head-level result is a
bounded null rather than proof of absence: at $n=60$/class we can exclude any single-head
effect larger than about $0.05$ AUC, not rule out a smaller one. Labels throughout are
subreddit membership, a noisy proxy for clinical status and a known construct-validity
limitation of social-media mental-health labels~\citep{chancellor2020methods}, and the
corpora are English-only.
A full accounting of these limitations, with sample sizes and margins, is in SI Section~\ref{sec:limitations}. We
neither claim the feature is the only route to the behavior, nor that ablating it is safe in
deployment; nothing here is a validated clinical instrument.

\section*{Methods}

\subsection*{The validation-gating protocol}
We use \emph{validation gating} for the discipline stated above: no mechanistic claim is
made about a behavior the model does not demonstrably perform, and every rung is reported
against a control that could have falsified it. Operationally, a concept enters the analysis
only once the model ranks its classes above a matched lexical baseline under the same prompt
(the behavioral gate). The rungs that follow, decodability, necessity, intrinsic dimension,
sufficiency and generality, are each paired with such a control: a magnitude-matched random
direction for necessity, a random orthonormal subspace for rank, an off-target readout for
sufficiency, a register-matched contrast for specificity. A rung whose measurement does not
separate from its control is reported as a negative rather than dropped, which is why four
of the eleven rows of Table~\ref{tab:summary} are negatives. The protocol is concept-agnostic;
SI Section~\ref{sec:secondconcept} runs it end to end on a second concept, toxicity.

\subsection*{Notation and readouts}

\begin{figure}[tbp]\centering
  \includegraphics[width=0.72\linewidth]{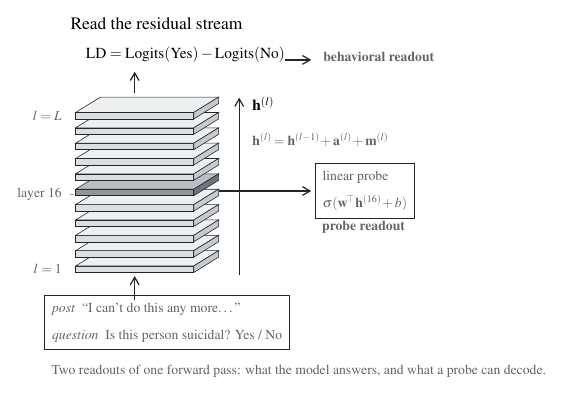}
  \caption{\textbf{Reading the residual stream.} Each post is posed as a yes/no question and
  read at the final token. Layers add into a shared stream (Equation~\eqref{eq:resid}), and a
  single forward pass yields two readouts of it: the model's own logit difference
  (Equation~\eqref{eq:ld}) and a linear probe on $\mathbf{h}^{(16)}$.}
  \label{fig:readout}
\end{figure}
We use the residual-stream notation of Elhage et al.~\citep{elhage2021framework} for the
transformer architecture of Vaswani et al.~\citep{vaswani2017attention}. Each of the
$L$ layers adds a multi-head attention output
$\mathbf{a}^{(l)}=\sum_{h=1}^{H}\mathbf{a}^{(l,h)}$ (a sum over $H$ heads) and an MLP output
$\mathbf{m}^{(l)}$ into a shared per-token residual stream,
\begin{equation}
\mathbf{h}^{(l)} = \mathbf{h}^{(l-1)} + \mathbf{a}^{(l)} + \mathbf{m}^{(l)},
\label{eq:resid}
\end{equation}
omitting layer normalization, which is not essential for the analyses
below~\citep{liu2025relevance}. The final representation is mapped to vocabulary logits by an
unembedding head, which intermediate states can also be read through directly, as in the
logit lens and its calibrated variants~\citep{nostalgebraist2020,belrose2023tunedlens},
\begin{equation}
\mathrm{Logits} = W_{U}\,\mathbf{h}^{(L)} + \mathbf{b} \in \mathbb{R}^{|\mathcal{V}|}.
\label{eq:logits}
\end{equation}
Because heads $\mathbf{a}^{(l,h)}$, MLPs $\mathbf{m}^{(l)}$, and the stream
$\mathbf{h}^{(l)}$ share one space, a single direction can be read out of $\mathbf{h}^{(l)}$
(probing), removed from it (ablation, Equation~\eqref{eq:ablate}), added to it (steering,
Equation~\eqref{eq:steer}), or have one summand swapped for its value under a different input
(patching, Equation~\eqref{eq:ie}).

Each task is posed as a yes/no question and read at the final token position (Fig.~\ref{fig:readout}). We define the
logit difference between the two answer tokens, taken from Equation~\eqref{eq:logits},
\begin{equation}
\begin{aligned}
&\mathrm{LD} = \mathrm{Logits}(\texttt{`` Yes''}) - \mathrm{Logits}(\texttt{`` No''}),\\
&P(\text{Yes}) = \sigma(\mathrm{LD}),
\end{aligned}
\label{eq:ld}
\end{equation}
where $\sigma$ is the logistic function. We use $\mathrm{LD}$, the readout of Liu et
al.~\citep{liu2025relevance} on this model family, so results are directly comparable to that
work's relevance-judgment circuit.

The representational readout is the residual-stream state at the answer-token position. For
decodability, a linear probe (standardized logistic regression, 5-fold cross-validation) is
fit per layer and scored by ROC-AUC. For the causal interventions we need a single fixed
vector rather than a probe hyperplane, so we define the suicidality direction at layer $l$ as
the unit-normalized difference of class means,
\begin{equation}
\mathbf{r}_l = \bar{\mathbf{h}}^{(l)}_{\text{suicide}}
            - \bar{\mathbf{h}}^{(l)}_{\text{non-suicide}},
\qquad
\hat{\mathbf{r}}_l = \mathbf{r}_l / \lVert \mathbf{r}_l \rVert,
\label{eq:dir}
\end{equation}
read at layer 16 for Llama. Layers are indexed by position in the residual-stream state
stack, where index~0 is the embedding output and index~$l$ is the output of decoder block
$l-1$; interventions hook the block feeding the indexed state. The direction is taken at
layer~16, adjacent to the probe's peak at layer~15, and the same index is used for every
causal intervention. We use difference-of-means rather than a probe
weight vector because it is training-free and recovers the behaviorally causal
direction~\citep{arditi2024refusal}.

\subsection*{Causal interventions}

\begin{figure}[tbp]\centering
  \includegraphics[width=0.66\linewidth]{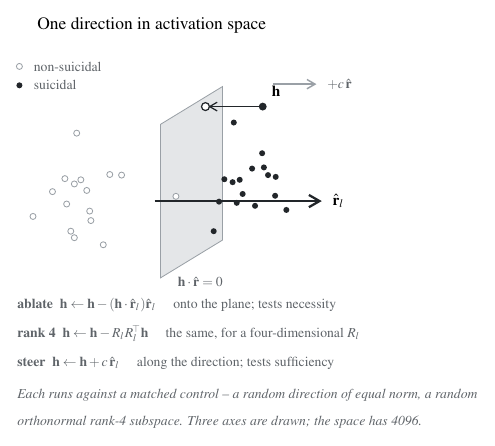}
  \caption{\textbf{The geometry of the interventions.} The two classes separate along
  $\hat{\mathbf{r}}_l$. Directional ablation (Equation~\eqref{eq:ablate}) projects the
  activation onto the hyperplane orthogonal to that direction; rank-$k$ ablation
  (Equation~\eqref{eq:rankk}) removes a $k$-dimensional $R_l$ in the same way; steering
  (Equation~\eqref{eq:steer}) slides along it. Three axes are drawn; the space has 4096.}
  \label{fig:geometry}
\end{figure}
Ablation tests necessity; steering tests sufficiency; patching localizes where the signal is
written; Fig.~\ref{fig:geometry} draws the geometry of the first three.

To test necessity, directional ablation removes the suicidality direction from the stream by
projecting it out at every token position, at one or all layers:
\begin{equation}
\mathbf{h}^{(l)} \leftarrow
\mathbf{h}^{(l)} - \bigl(\mathbf{h}^{(l)}\!\cdot\hat{\mathbf{r}}_l\bigr)\hat{\mathbf{r}}_l.
\label{eq:ablate}
\end{equation}
If $\hat{\mathbf{r}}_l$ carries the behavior, Equation~\eqref{eq:ablate} should collapse
$\mathrm{LD}$ and the suicide-vs-non-suicide AUC; an equal-norm random direction should not.
Control: a random unit direction in place of $\hat{\mathbf{r}}_l$.

To estimate intrinsic dimension we build a rank-$k$ subspace per layer by iterative
deflation: take the difference-of-means direction, project it out of the activations,
recompute the difference of means on the residual, and repeat $k$ times; orthonormalize the
result into $R_l=[\mathbf{u}_1,\dots,\mathbf{u}_k]$ and project the whole subspace out,
\begin{equation}
\mathbf{h}^{(l)} \leftarrow \mathbf{h}^{(l)} - R_l R_l^{\top}\mathbf{h}^{(l)}.
\label{eq:rankk}
\end{equation}
The smallest $k$ at which AUC reaches chance would estimate the dimensionality of the
behavioral subspace if the sweep were monotone. It is not: AUC falls below chance at some
ranks, which marks a rotated decision axis rather than an erased one, and a sign-blind
threshold cannot tell the two apart. We therefore report the sweep and treat the dimension
as unidentified. Control: a random orthonormal rank-$k$ subspace.

Steering tests the converse, whether adding the feature induces the behavior, following
activation addition~\citep{turner2023actadd}, by injecting it
into the residual stream of non-suicidal text across layers $12$--$20$:
\begin{equation}
\begin{aligned}
\mathbf{h}^{(l)} &\leftarrow \mathbf{h}^{(l)} + c\,\Delta_l\,\hat{\mathbf{r}}_l,\\
\Delta_l &= \bigl(\bar{\mathbf{h}}^{(l)}_{\text{suicide}}
                - \bar{\mathbf{h}}^{(l)}_{\text{non-suicide}}\bigr)\!\cdot\hat{\mathbf{r}}_l,
\end{aligned}
\label{eq:steer}
\end{equation}
where $\Delta_l$ is the class-mean projection gap at layer $l$, so $c=1$ moves a post by one
class-mean separation. Controls: a magnitude-matched random direction, and an off-target
yes/no readout scored under the same perturbation.

Activation patching localizes where in the network the signal is written, by transplanting
one component's activation between two
inputs~\citep{meng2022rome,heimersheim2024patching}. We use the denoising variant on
contrastive pairs: a clean (suicidal) post $X_{\text{clean}}$, whose expected answer is
``Yes,'' and a corrupt (non-suicidal) post $X_{\text{corrupt}}$, whose expected answer is
``No.'' For a chosen component $z\in\{\mathbf{a}^{(l)},\mathbf{m}^{(l)}\}$ we run the model
three times: a clean run caching $z$, a corrupt run recording
$\mathrm{LD}_{\text{corrupt}}$, and a patched run on $X_{\text{corrupt}}$ in which $z$ alone
is restored to its clean value, recording $\mathrm{LD}_{\text{patched}}$. The causal
contribution of $z$ is the normalized indirect
effect~\citep{wang2023ioi,liu2025relevance}
\begin{equation}
\mathrm{IE} =
\frac{\mathrm{LD}_{\text{patched}} - \mathrm{LD}_{\text{corrupt}}}
     {\mathrm{LD}_{\text{clean}} - \mathrm{LD}_{\text{corrupt}}}.
\label{eq:ie}
\end{equation}
The normalization fixes $\mathrm{IE}\approx1$ when restoring $z$ alone recovers the clean
behavior, $\mathrm{IE}\approx0$ when it does nothing, and $\mathrm{IE}<0$ when it worsens the
corrupt baseline. The same metric, applied per attention head by patching
$\mathbf{a}^{(l,h)}$ or by zeroing a head's contribution (knockout), drives the head-level
localization reported in SI Section~\ref{sec:mechfigs}.

Every intervention is paired with a matched control: a random direction or subspace of equal
magnitude (ablation, steering), or an off-target yes/no readout (steering selectivity). An
effect is therefore attributable to the direction and its concept match, not to the
perturbation's norm alone.

\subsection*{Models and data}
Primary model: Llama-3.1-8B-Instruct (32 layers, 32 heads) in fp16 on a single 96\,GB GPU;
SI Table~\ref{tab:precision} reports the same rungs under 4-bit NF4 quantization. Cross-family replication uses Qwen2.5-7B-Instruct and
Mistral-7B-Instruct-v0.3/v0.1; the cross-scale sweep adds Llama-3.2-1B/3B-Instruct and
Qwen2.5-0.5B/1.5B/3B/7B-Instruct. Five existing corpora are used as released, with no new data
collection: Suicide-Watch-232K~\citep{komati2021suicidewatch} (primary feature analysis),
SDCNL~\citep{haque2021sdcnl} and SWMH~\citep{ji2022mentalbert} (generalization; SWMH also
supplies a separable depression class for the specificity test), the synthetic
DeepSuiMind~\citep{li2025implicit} (implicit-intent task), and the access-controlled UMD
Reddit Suicidality Dataset v2~\citep{shing2018umd,zirikly2019clpsych} (used only for the
severity validation). UMD assigns one ordinal risk level per \emph{user}, not per post,
while the corpus itself is longitudinal; we therefore make the user the unit of analysis,
concatenating each user's SuicideWatch posts most-recent-first and truncating at 4,000
characters. Of the 490 annotated users, the 245 with SuicideWatch posts enter the analysis
($9\%$ reach the truncation cap). Because posting volume itself rises with severity, the
severity result is reported with length partialled out. A 5-fold TF-IDF +
logistic-regression model, scored as an AUC on the same examples and folds as the model
readout, supplies the lexical baseline throughout. Full dataset roles are in SI
Table~\ref{tab:datasets} and quantization/robustness details in SI
Section~\ref{sec:precision}.

All five corpora are used under their stated terms for non-commercial research, with no
redistribution of raw text and no attempt at re-identification. Three are de-identified
public Reddit text: Suicide-Watch-232K~\citep{komati2021suicidewatch} (Kaggle, CC~BY-SA~4.0);
SWMH~\citep{ji2022mentalbert} (CC~BY-NC~4.0); and SDCNL~\citep{haque2021sdcnl} (released for
research). DeepSuiMind~\citep{li2025implicit} is synthetic dialogue. The UMD Reddit
Suicidality Dataset v2~\citep{shing2018umd,zirikly2019clpsych} is access-controlled and was
obtained under its data-usage agreement following governance-committee review; we report only
aggregate statistics from it, never post text.

\subsection*{Prompt templates}
All three templates share the same body slot (\texttt{\{text\}}, the post) and the same
yes/no question; they differ only in the system framing. The body is truncated to fit a
256-token budget by trimming the body, never the question or the \texttt{Answer:} suffix.
Verbatim:

\begin{promptlabel}{standard}
\begin{lstlisting}[style=prompt]
Read the following text and determine if the person shows signs of suicidal ideation.

Text: {text}

Does this text show signs of suicidal ideation? Answer with Yes or No.
Answer:
\end{lstlisting}
\end{promptlabel}

\begin{promptlabel}{empathetic}
\begin{lstlisting}[style=prompt]
You are a compassionate mental health professional. Read the following text
carefully, paying attention to emotional undertones and implicit cues of distress.

Text: {text}

Does this text show signs of suicidal ideation? Answer with Yes or No.
Answer:
\end{lstlisting}
\end{promptlabel}

\begin{promptlabel}{clinical}
\begin{lstlisting}[style=prompt]
You are a clinical psychologist conducting a risk assessment. Analyze the
following text for any indicators of suicidal ideation, including implicit
associations with death.

Text: {text}

Does this text show signs of suicidal ideation? Answer with Yes or No.
Answer:
\end{lstlisting}
\end{promptlabel}

All binary-task experiments use \emph{standard}. The implicit-intent NO-GO was checked
against all three templates and both answer-token conventions (\texttt{`` Yes''} and
\texttt{``Yes''}); none cleared the lexical baseline. Worked input-to-output examples, and
the robustness of the readout to template choice, are in SI Section~\ref{sec:ethics}.

\subsection*{Statistics and reproducibility}
Tests of difference (DeLong, paired bootstrap, Spearman) are two-tailed and evaluated at
$\alpha=0.05$; the TOST equivalence test comprises two one-sided tests, each at
$\alpha=0.05$. Exact $p$-values are reported throughout rather than thresholds. Headline AUCs carry 2,000-sample bootstrap 95\% confidence
intervals, which make no distributional assumption and are used in preference to parametric
intervals because AUCs are bounded and their sampling distribution is skewed near the
ceiling. Margins over the lexical baseline are tested with DeLong's test for correlated ROC
curves~\citep{delong1988}, which is the appropriate test when two scores are computed on the same examples, plus
a stratified case-resampled paired bootstrap of the AUC difference (10,000 resamples).
Severity readouts are compared with a two-one-sided-tests (TOST) equivalence test, since the
claim there is equivalence rather than difference, on Fisher-$z$ transformed correlations with
an equivalence margin of $\Delta r = 0.15$; because per-user readout pairs were not cached we
use the independent-samples standard error, which overstates the variance and is therefore
conservative for an equivalence test. Expert-severity association uses Spearman's $\rho$, a
rank statistic chosen because the clinician labels are ordinal and not assumed
interval-scaled, with post length partialled out. Per-comparison sample sizes are given with each result in the text and in
full in SI Table~\ref{tab:summary-full}.

Reported $p$-values for the seven positive rungs (behavioral gate, decodability,
suicide-vs-depression specificity, necessity, cross-dataset transfer to SDCNL and SWMH, and
the expert-severity correlation) carry a Benjamini--Hochberg correction across the family
(largest $q=0.04$); the two null results (absence of a critical head, failure of
concept-selective steering) are reported as bounded nulls and excluded from that family.

All reported experiments run in fp16 on a single 96\,GB GPU. The quantized replications in SI
Table~\ref{tab:precision} run the same scripts under a single environment variable, with
weight folding and centering disabled, since these corrupt quantized weights. Directions are
fit on training splits and evaluated held-out throughout; where a threshold is used it is
calibrated on the training split alone. Every row of the evidence table is produced by a
dedicated script named for the claim and written to a corresponding result file, with the
exact command, prompt template, output path, model versions, precision setting, dependency
versions, and random seeds recorded alongside it, and every figure rendered from those result
files.

\section*{Data availability}
Four of the five corpora are openly available from the original sources cited in Methods,
under their stated terms: the de-identified public Reddit datasets
Suicide-Watch-232K~\citep{komati2021suicidewatch}, SDCNL~\citep{haque2021sdcnl} and
SWMH~\citep{ji2022mentalbert}, and the synthetic dialogue corpus
DeepSuiMind~\citep{li2025implicit}. The fifth, the UMD Reddit Suicidality Dataset
v2~\citep{shing2018umd,zirikly2019clpsych}, is access-controlled and obtainable only under
its data-usage agreement. Fine-grained model
artifacts (probe weights, SAE dictionaries, steering vectors) are available to vetted
researchers on request, gated in line with safe-messaging practice (SI Section~\ref{sec:ethics}).

\section*{Code availability}
All code, configurations, and per-claim result files needed to regenerate every number,
table, and figure are released at
\url{https://github.com/INQUIRELAB/mech-interp-on-suicide}, with the exact command,
prompt template, output path, and environment documented per result (Methods, ``Statistics
and reproducibility'').

\section*{Acknowledgements}
We thank the American Association of Suicidology for access to the UMD Reddit Suicidality
Dataset under its data-usage agreement. This research received no specific grant from any
funding agency in the public, commercial, or not-for-profit sectors.

\section*{Author contributions}
N.A.\ conceived the study, designed and implemented the framework and all experiments,
analysed the results, produced the figures, and drafted the manuscript. D.S.\ validated the
clinical-severity data and the expert risk annotations. M.B.\ and S.S.\ supervised the work,
and M.B.\ secured access to the UMD dataset. All authors reviewed and edited the manuscript
and approved the final version.

\section*{Competing interests}
The authors declare no competing interests.

\section*{Ethics declaration}
This work studies how a model represents suicidal content to audit model behavior;
it is not a clinical instrument and we release no screening tool. We use five existing
corpora with no new data collection or human-subjects recruitment; the UMD dataset was
obtained under its data-usage agreement following governance-committee review, and we report
only aggregate statistics, never post text. We follow safe-messaging guidance throughout
(we neither generate nor describe means of self-harm) and gate fine-grained model artifacts
to vetted researchers rather than releasing them openly, since the same direction could steer
a model toward distress-affirming output. Nothing here is validated for clinical use. Readers
in distress can reach the 988 Suicide and Crisis Lifeline (US) or
\href{https://findahelpline.com}{findahelpline.com}. Full ethics and data-provenance detail
is in SI Section~\ref{sec:ethics}.

\section*{Supplementary information}
The appendix (Supplementary Information) contains extended methods and
statistics, the full localization analysis (per-head, per-edge, and sparse-autoencoder
results), prompt templates and worked examples, additional robustness checks (8-bit
replication, small-model positive controls), and the complete evidence table with every
confidence interval.

\clearpage
\appendix
\setcounter{figure}{0}\setcounter{table}{0}\setcounter{equation}{0}\setcounter{section}{0}
\renewcommand{\thefigure}{S\arabic{figure}}
\renewcommand{\thetable}{S\arabic{table}}
\renewcommand{\theequation}{S\arabic{equation}}
\renewcommand{\thesection}{S\arabic{section}}
\begin{center}{\Large\bfseries Supplementary Information}\end{center}
\bigskip
\section{Localization: full figures and edge-level analysis}
\label{sec:mechfigs}
The figures and per-head statistics behind the main-text summary (``No sparse head- or
edge-level circuit'') follow; all use the validated binary task with the same model,
prompt, and readout as the main experiments. Per-head knockout (zeroing each head,
$n=60$/class) finds no single critical head: the largest effect is a $0.013$ AUC drop with a
bootstrap CI spanning zero ($[-0.019,0.051]$), no head survives Benjamini--Hochberg
correction ($0$ of $1{,}024$ at $q<0.05$), and the mean effect across all heads is
$+0.0001$. The bound is therefore that no single head accounts for more than about $0.05$
AUC. A smaller $n=20$/class run gives the same null with a looser bound (largest drop
$0.055$, CI $[-0.055,0.188]$) but nominates a different top head (L11H1 rather than L9H17),
and that instability across sample size is itself what one expects when no head carries the
behavior. Sample sizes here match between the sweep and the bootstrap statistics; they must,
because the statistics re-use the sweep's examples by seed.

\begin{figure*}[tb]\centering
  \figorbox{figures/fig6_head_knockout.png}
  \caption{Per-head knockout importance across all $32\times32$ attention heads
  (binary task, $n=20$/class). Left: AUC drop from zeroing each head;
  importance concentrates in mid layers (peak around L11--L12) and falls to near
  zero from L18 onward, consistent with a mid-layer write that is read out at L15,
  with no single dominant head. Right: the same drop summarized by depth
  (max and mean over heads in each layer).}
  \label{fig:headmap}
\end{figure*}

A circuit claim is a claim about edges: that a small set of upstream$\to$downstream
connections routes the behavior. We test this with edge attribution
patching~\citep{syed2023attribution}, scoring all $2145$ causal edges of the
residual-stream graph in one forward+backward pass per contrastive pair ($n=28$),
normalized by the clean$-$corrupt logit-difference gap as in Equation~\eqref{eq:ie}. To
guard against the linearization error of attribution patching, we then run the exact
single-edge path patch on the top-ranked edges. The picture is broadly distributed
(Figure~\ref{fig:eap}): attribution is spread near the uniform-null level (the five
strongest edges hold $3.6\%$ of total $|$attribution$|$ versus $0.2\%$ under a uniform
null, the top ten hold $5.9\%$, and $641$ of $2145$ edges, roughly a third, are needed to
reach $80\%$; Gini $0.63$). Only the trivial direct-to-logit write survives exact
path patching (mlp30$\to$logits, EAP $-0.56$ and exact IE $-0.56$), whereas high-scoring
internal routing edges collapse under exact verification (mlp14$\to$attn24 EAP $+0.84$ but
exact $-0.08$; mlp30$\to$attn31 exact $-0.01$): the twelve strongest edges together recover
only $+0.27$ of the clean$-$corrupt gap, and EAP-vs-exact rank agreement is modest
(Spearman $0.33$). No small, faithful edge sub-circuit exists.

\begin{figure*}[tb]\centering
  \figorbox{figures/edge_attribution.png}
  \caption{Edge attribution patching (EAP) finds no sparse sub-circuit.
  Top: EAP edge-attribution matrix over the $2145$ causal edges
  (upper-triangular; upstream writes $\times$ downstream reads). Middle:
  cumulative $|$attribution$|$ ranked by edge magnitude (dark) against the
  uniform-null line (grey); the curve hugs the null, with $\sim\!698$ edges needed for
  $80\%$ (Gini $0.63$). Bottom: EAP score vs.\ exact single-edge path-patch
  indirect effect for the top-12 edges; only the direct-to-logit output write lies
  near the diagonal, while high-EAP internal edges collapse to near-zero exact effect
  (Spearman $0.33$; top-12 recover only $0.27$ of the clean$-$corrupt gap).}
  \label{fig:eap}
\end{figure*}

\begin{figure*}[tb]\centering
  \figorbox{figures/sae_features_L16.png}
  \caption{Sparse autoencoder at L16 (binary task). Top: reconstruction and
  L1 loss over training. Middle: per-feature mean activation on suicide
  vs.\ non-suicide posts; the top suicide-selective atoms (red) lie far above the
  diagonal. Bottom: injecting the top-10 features into the L16 residual of
  non-suicidal posts shifts the Yes$-$No logit difference rightward, past the class
  midpoint: the feature-level analogue of the rank-$k$ steering result, subject to
  the same specificity caveat. This is the one analysis in the paper still measured
  under 4-bit quantization; it is illustrative rather than load-bearing, and no
  claim in the main text depends on it.}
  \label{fig:sae}
\end{figure*}

\begin{figure*}[tb]\centering
  \figorbox{figures/patching_ie.png}
  \caption{Layer-wise denoising activation patching (binary task, 19 clean/corrupt
  pairs). Indirect effect of restoring each layer's attention (red) or MLP (blue)
  contribution to the residual stream; both peak at L11--L12 and fall to near zero
  by L20, locating the suicidality write in the mid layers.}
  \label{fig:patching}
\end{figure*}

\begin{figure*}[tb]\centering
  \figorbox{figures/dla.png}
  \caption{Direct logit attribution. Per-layer contribution to the Yes$-$No
  logit difference (suicidal minus non-suicidal posts) for attention and MLP. The
  decision logit is expressed by mid-to-late components (L13--24), downstream of the
  mid-network write; the decomposition reconstructs the true logit difference to
  $3\times10^{-3}$ logits.}
  \label{fig:dla}
\end{figure*}

\subsection{Activation patching and the indirect effect}
\label{sec:patching}
The denoising patching procedure and the normalized indirect effect $\mathrm{IE}$ used
throughout this section are defined in the Methods of the main text. Recall that
$\mathrm{IE}\approx1$ when restoring a component alone recovers the clean behavior,
$\mathrm{IE}\approx0$ when it does nothing, and $\mathrm{IE}<0$ when it worsens the corrupt
baseline. The same metric, applied per attention head by patching $\mathbf{a}^{(l,h)}$ or by
zeroing a head's contribution (knockout), drives the head-level localization above.

\section{The framework transfers to a second concept, and downstream detection}
\label{sec:secondconcept}
To support the framework claim beyond suicidality, we run the complete
gate$\to$decodability$\to$necessity$\to$rank ladder on toxicity (civil\_comments) with the
same code, changing only the two classes and the yes/no question. Toxicity behaves like
suicidality: the gate passes (behavioral AUC $0.896$, probe $0.925$ at layer 14, above the
$0.802$ lexical baseline), the direction is causally necessary (ablation drops AUC from
$0.910$ to $0.567$, random direction inert at $0.913$), and the behavior is low-rank
($k^\ast=1$). We count this as a distinct concept; the suicide-vs-depression run in the main
text is the same concept under a harder negative, a specificity control rather than a second
concept.

\begin{table}[tb]\centering
\begin{tabular}{lcccc}
\toprule
condition & AUC & F1 & recall & F1 drop \\
\midrule
Full model & $0.975$ & $0.918$ & $0.890$ & -- \\
Random-direction ablation & $0.975$ & $0.918$ & $0.890$ & $+0.000$ \\
Suicidality-direction ablation & $0.832$ & $0.690$ & $0.590$ & $+0.228^{*}$ \\
Rank-4 subspace ablation & $0.766$ & $0.000$ & $0.000$ & $+0.918^{*}$ \\
\bottomrule
\end{tabular}
\caption{Downstream suicide-detection metrics ($n=100$/class test set, fixed
train-calibrated threshold). Direction and threshold are fit on a disjoint
$n=100$/class train split and scored on this independent test split, so absolute AUCs
differ from the faithfulness eval in the main text (which measures suicide-vs-non-suicide
ranking on its own set); the pattern is identical. Ablating the suicidality feature
destroys detection while an equal-norm random direction does nothing. $^{*}$F1 drop vs.\
full model significant at $p<0.05$ (paired bootstrap). Analogue of Liu et al.'s Table~2
component knockout~\citep{liu2025relevance}.}
\label{tab:downstream}
\end{table}

\section{Extended limitations}
\label{sec:limitations}
The model's strong affirmative bias forces us onto ranking (AUC) and projections rather than
accuracy. The binary task is partly lexical (matched-sample TF-IDF AUC $0.922$); our
contribution is the margin over lexical cues, significant for both the behavioral gate and
the probe but small on 232K, and the register-matched suicide-vs-depression separation is
modest (probe $0.80$ clears its lexical baseline, $p=0.04$; the behavioral readout $0.74$
does not alone). The binary labels are subreddit membership, a noisy proxy that also carries
register and topic differences, so the headline AUCs reflect a population contrast, not a
clean clinical one; the matched-negative and keyword-masking results net out the proxy, and
the decisive claims are scoped accordingly.

Several findings are bounded in generality. On expert-annotated UMD data the internal
projection ($\rho=0.56$) does not beat the behavioral readout on the full sample ($\rho=0.56$;
the two are statistically equivalent, TOST $p=0.01$); the internal readout's advantage
appears only on the subset the model itself gets wrong (Section~\ref{sec:monitor}), and the
finer severity resolution is shown on text-type proxies rather than the expert risk levels.
The behavioral readout holds only from
$\approx$1.5B up (at $\approx$1B it is at chance though the representation is strong), and
clean single-direction necessity is uneven below the 7--8B class (e.g.\ Qwen2.5-3B
$0.94\to0.89$ is near-inert, needing the rank-$k$ test), so we establish the central
low-rank necessity result on the primary model; the datasets are Reddit-derived and
English-only. The encoding-vs-use dissociation below $\approx$3B is supported by a positive
control (small models perform other yes/no tasks but not suicidality), with the residual
risk that the control tasks are easier than suicidality for reasons beyond use.

Other caveats are rung-specific. Steering establishes only a magnitude-controlled effect,
not concept-selective sufficiency, and saturates at high $c$; necessity, low-rank, and
transfer do not involve steering. Head-level attribution is a bounded null at $n=60$/class:
it excludes any single-head effect above $\approx$0.05 AUC rather than proving no head-level
structure. All mechanistic results are measured in fp16 on a single 96\,GB GPU. Because an
earlier version of this work ran on a 12\,GB card, we also hold 4-bit NF4 and 8-bit
replications, reported in Table~\ref{tab:precision}: every qualitative conclusion is
precision-robust, but 4-bit \emph{overstates} single-direction necessity (ablated AUC $0.66$
against $0.83$ in fp16). The rank-$k$ entries in that table come from the original sparse
sweep and are superseded by the dense replication in Section~\ref{sec:ranksweep}, which shows
the rank of the necessary subspace is not identifiable; we therefore locate the causal claim
at the subspace without asserting its dimension.
Several principal comparisons rest on modest samples ($n=150$/class cross-model,
$n=100$/class downstream, $n=150$ register-matched, $n=245$ UMD users, $n=60$ heads, $n=28$
edges); we report bootstrap confidence intervals throughout, but the strong-sounding margins
should be read against these sample sizes.

\section{Internal monitor versus the model's own answer}
\label{sec:monitor}
The encoding-versus-use dissociation is reported across scale in the main text. It
also has a within-model consequence at 8B, which we test as follows. A probe and a
difference-of-means direction are fit on $500$ posts per class of 232K at the peak
layer; one decision threshold per readout is calibrated on a disjoint $200$ per class;
and both are scored on a further $1{,}000$ per class. Every readout sees the same
forward passes, so the comparison is not confounded by input differences.

The model's behavioral readout reaches AUC $0.961$ at recall $0.796$, missing $204$ of
the $1{,}000$ suicidal posts. The probe reaches AUC $0.973$ at recall $0.930$, missing
$70$. At its own calibrated threshold the probe places $73\%$ of the model's $204$ missed
posts above the line, but that threshold sits at higher recall, and a readout that flags more
posts recovers misses for free. The comparison must therefore hold the operating point fixed,
which we do in two ways (Table~\ref{tab:matched}).

Matching the alarm rate is the strictest test: constrained to flag exactly the $839$ posts
the model flags, the probe attains higher recall \emph{and} higher precision, and $45.6\%$ of
the model's misses fall inside its alarm set. It is choosing a different set of the same
size, not a larger one. Matching precision instead at $0.949$ gives recall $0.872$ against
$0.796$ and recovers $57.4\%$. The recovery figure therefore ranges from $46\%$ to $73\%$
depending on how the operating point is fixed; we quote the matched-alarm-rate figure in the
main text as the most conservative. The effect transfers without refitting: $27\%$ of the
model's misses are recovered on SDCNL and $30\%$ on SWMH. Per-example held-out scores for all
three readouts are released with the result file, so any other operating point can be checked
without re-running the model.

\begin{table}[htb]\centering\footnotesize
\begin{tabular}{llrrrr}
\toprule
readout & operating point & precision & recall & flagged & FN recovered \\
\midrule
behavioral (model) & own threshold & $0.949$ & $0.796$ & $839$ & -- \\
probe & own threshold & $0.919$ & $0.930$ & $1011$ & $0.730$ \\
probe & matched precision & $0.949$ & $0.872$ & $919$ & $0.574$ \\
probe & matched alarm rate & $0.962$ & $0.807$ & $839$ & $0.456$ \\
projection & matched precision & $0.949$ & $0.760$ & $801$ & $0.083$ \\
projection & matched alarm rate & $0.944$ & $0.792$ & $839$ & $0.142$ \\
\bottomrule
\end{tabular}
\caption{The internal readout's advantage under matched operating points
($n=1{,}000$/class held out). ``FN recovered'' is the fraction of the $204$ suicidal posts
the model itself misses that the readout flags. The probe beats the behavioral readout on
both precision and recall at an identical alarm rate; the difference-of-means projection
does not.}
\label{tab:matched}
\end{table}

Three limits belong with the claim. First, the advantage is carried by the trained probe, not
by the difference-of-means direction that supports the causal results: at a matched alarm
rate the projection recovers only $14\%$ of the misses and its precision is no better than
the model's. The readout that is most useful for monitoring is therefore not the one that
carries the causal claim, and the two should not be conflated. Second, within the set of
cases the model gets wrong, the probe cannot rank them (AUC $0.52$), so the gain is a better
operating point rather than better discrimination on the hard cases. Third, this is a
monitoring result on static posts under a fixed prompt; it is not a validated clinical
instrument, and we
make no claim about deployment.

\section{Dense rank-$k$ replication}
\label{sec:ranksweep}

The dimensionality claim in earlier drafts of this work rested on a sparse sweep,
$k\in\{1,2,4,8\}$, measured at $n=80$/class from a single seed with no interval, and scored
by taking the $k$ of minimum AUC. That design cannot settle the question: $k=3$ is never
measured, so ``four'' is not distinguishable from three; a single seed gives no handle on
variance; and the argmin estimator is not the estimator the Methods define.

We therefore re-ran the sweep under a decision rule fixed before the run: $k^\ast$ is the
smallest $k$ whose bootstrap 95\% CI lower bound falls at or below $0.5$, taken by majority
across seeds, with the pre-specified outcomes that $k^\ast=4$ would leave the original claim
standing, another single value would replace it, and disagreement across seeds would force a
retreat to ``low-dimensional''. The replication is dense and gapless through $k=10$ (plus
$12$ and $16$), in fp16, at $n=150$ train and $n=200$ test per class, over three seeds, with
2{,}000-sample bootstrap intervals at every rank.

Table~\ref{tab:ranksweep} gives the result. Two things hold. Ablation degrades behavior at
every rank, from a baseline of $0.963$ to between $0.53$ and $0.77$, and the random rank-$k$
control is inert everywhere, spanning only $0.950$--$0.967$ across all ranks and all seeds.
Two things fail. The curve is not monotone and does not settle at chance: per-seed $k^\ast$
comes out $3$, $3$ and $10$, the seed-to-seed spread reaches $0.38$ AUC at $k=16$ and $0.32$
at $k=3$, and rank-4 in particular sits at $0.751$ averaged over seeds (individually $0.754$,
$0.753$, $0.745$), not the $0.515$ of the original sweep. And at four points the ablated AUC
falls below chance, as low as $0.331$; an AUC beneath $0.5$ means the ranking has inverted,
so the information is still present and the decision axis has been rotated rather than
erased. A sign-blind threshold on AUC cannot separate that case from genuine removal, which
is what the pre-registered estimator does.

We therefore take the third pre-specified branch. Necessity is a property of a
low-dimensional subspace rather than of any single direction, and the dimension of that
subspace is not identified by rank-$k$ ablation scored this way. Recovering it would need an
estimator that is invariant to sign, a subspace construction whose fitted directions are
stable across resamples, or both.

\begin{table}[htb]\centering\footnotesize
\begin{tabular}{lcccr}
\toprule
$k$ & seed 0 & seed 1 & seed 2 & random \\
\midrule
1 & 0.780 [0.73, 0.82] & 0.696 [0.64, 0.74] & 0.718 [0.67, 0.77] & 0.962 \\
2 & 0.610 [0.55, 0.66] & 0.693 [0.64, 0.74] & 0.594 [0.54, 0.65] & 0.962 \\
3 & 0.356 [0.30, 0.41] & 0.544 [0.49, 0.60] & 0.677 [0.62, 0.73] & 0.963 \\
4 & 0.754 [0.70, 0.80] & 0.753 [0.71, 0.80] & 0.745 [0.70, 0.79] & 0.962 \\
5 & 0.806 [0.76, 0.85] & 0.654 [0.60, 0.70] & 0.695 [0.64, 0.75] & 0.962 \\
6 & 0.838 [0.80, 0.88] & 0.667 [0.61, 0.72] & 0.804 [0.76, 0.85] & 0.962 \\
7 & 0.872 [0.84, 0.90] & 0.592 [0.54, 0.65] & 0.794 [0.75, 0.84] & 0.963 \\
8 & 0.822 [0.78, 0.86] & 0.561 [0.51, 0.62] & 0.675 [0.62, 0.73] & 0.963 \\
9 & 0.630 [0.57, 0.69] & 0.612 [0.56, 0.67] & 0.675 [0.62, 0.72] & 0.963 \\
10 & 0.444 [0.39, 0.50] & 0.703 [0.65, 0.75] & 0.550 [0.49, 0.60] & 0.962 \\
12 & 0.480 [0.43, 0.54] & 0.571 [0.51, 0.63] & 0.634 [0.58, 0.69] & 0.962 \\
16 & 0.331 [0.28, 0.38] & 0.580 [0.52, 0.64] & 0.711 [0.66, 0.76] & 0.960 \\
\bottomrule
\end{tabular}
\caption{Dense rank-$k$ ablation, fp16, $n=200$/class test, with 2{,}000-sample bootstrap
95\% intervals. Baselines are $0.968$, $0.963$ and $0.958$ for seeds 0--2. The final column
is the random rank-$k$ control averaged over seeds. Per-seed $k^\ast=3,3,10$.}
\label{tab:ranksweep}
\end{table}

\section{Precision robustness}
\label{sec:precision}
Every reported number is measured in fp16. Because an earlier version of this
work was constrained to a 12\,GB card, we re-ran the identical scripts under 4-bit
NF4 quantization, switching precision through a single environment variable so that
sampling, seeds, and splits are held fixed. Every qualitative conclusion is
precision-robust. The two quantities that move are both ablation depths, and both
move in the same direction: 4-bit makes the causal interventions look stronger than
they are. Single-direction ablation reaches AUC $0.66$ in 4-bit against $0.83$ in
fp16. The rank-4 rows below ($0.40$ in 4-bit against $0.52$ in fp16) come from the
original single-seed sweep at $n=80$/class and do not survive replication: the dense
sweep of Section~\ref{sec:ranksweep} puts rank-4 at $0.751$ averaged over three seeds in
fp16. We therefore draw no precision conclusion from the rank rows, and locate the causal
claim at the subspace rather than at any single direction on the strength of the
single-direction comparison alone.

\begin{table}[htb]\centering\footnotesize
\begin{tabular}{lrrr}
\toprule
Quantity & 4-bit & fp16 & $\Delta$ \\
\midrule
Behavioral AUC (gate) & 0.951 & 0.961 & +0.011 \\
Probe AUC (peak layer) & 0.968 & 0.969 & +0.001 \\
TF-IDF baseline & 0.894 & 0.894 & +0.000 \\
Probe recall, masked & 0.745 & 0.740 & -0.005 \\
Necessity: baseline AUC & 0.970 & 0.975 & +0.005 \\
Necessity: ablated AUC & 0.709 & 0.832 & +0.122 \\
Necessity: random-dir AUC & 0.970 & 0.975 & +0.005 \\
Rank-4 ablated AUC & 0.402 & 0.515 & +0.113 \\
Rank-4 random AUC & 0.953 & 0.962 & +0.010 \\
Steering flip rate @ c=1 & 1.000 & 1.000 & +0.000 \\
Off-target LD @ c=1 & 12.841 & 13.109 & +0.268 \\
Transfer AUC (SDCNL) & 0.961 & 0.963 & +0.002 \\
Transfer AUC (SWMH) & 0.917 & 0.917 & +0.001 \\
\bottomrule
\end{tabular}
\caption{Headline quantities under 4-bit NF4 and fp16, same scripts and seeds.
Deltas above $0.05$ appear only in the two ablation depths; both indicate that
quantization exaggerates the measured causal effect.}
\label{tab:precision}
\end{table}

\section{Ethics, data provenance, and worked examples}
\label{sec:ethics}
Experimental procedure, prompt templates, environment, and the statistical protocol are
reported in the Methods of the main text; this section covers ethics and data provenance, and
gives worked input-to-output examples.

This work studies how a model represents suicidal content to audit model behavior; it is not
a clinical instrument and we release no screening tool. ``Deployed detector,'' ``downstream
classifier,'' and ``auditing'' refer to a model-development setting (probing or monitoring an
existing system's internal reliance on suicidality cues), not a patient-facing diagnostic. We
use five corpora and add no new data collection or annotation. Three are de-identified public
Reddit text: Suicide-Watch-232K~\citep{komati2021suicidewatch} (Kaggle, CC~BY-SA~4.0);
SWMH~\citep{ji2022mentalbert} (CC~BY-NC~4.0); and SDCNL~\citep{haque2021sdcnl} (released for
research). DeepSuiMind~\citep{li2025implicit} is synthetic dialogue. The access-controlled
University of Maryland Reddit Suicidality Dataset, Version 2~\citep{shing2018umd,
zirikly2019clpsych}, used only for the expert-severity validation, was obtained under its
data-usage agreement following governance-committee review; we report only aggregate
statistics, never post text. We use each corpus under its stated terms for non-commercial
research, do not redistribute raw text or attempt re-identification, and the worked examples
below are truncated and paraphrased. We report only classification behavior and internal
representations, in line with safe-messaging guidance, and neither generate nor describe
means of self-harm.

\subsection{Worked examples (input $\to$ output)}
\label{app:examples}
The body text is rendered through the standard template (Methods) and the chat template; we
report the model's argmax answer, the logit difference $\mathrm{LD}$ and $P(\text{Yes})$ as
defined in the Methods. Table~\ref{tab:examples}
shows a representative 4 of the $n=8$ per class on which the summary statistics are
computed (the full set and rendered prompts are released with the code).

\begin{table}[tb]\centering
\footnotesize\setlength{\tabcolsep}{4pt}\renewcommand{\arraystretch}{0.92}
\begin{tabular}{llrrp{0.42\linewidth}}
\toprule
true & pred & LD & $P(\text{Yes})$ & body (truncated) \\
\midrule
suicide & Yes & $7.60$ & $1.000$ & everything ruined\dots boyfriend broke\dots not want anything \\
suicide & Yes & $9.01$ & $1.000$ & really overwhelming\dots not know anymore\dots voices \\
suicide & Yes & $10.48$ & $1.000$ & going today\dots not sure really writing\dots \\
suicide & Yes & $3.54$ & $0.972$ & hope help content\dots \\
\midrule
non-suicide & Yes & $0.78$ & $0.686$ & anyone else server found someone precious\dots grateful \\
non-suicide & Yes & $1.34$ & $0.792$ & work yesterday girl asked number\dots \\
non-suicide & Yes & $6.58$ & $0.999$ & dad drug\dots guess sad dad\dots \\
non-suicide & Yes & $2.28$ & $0.907$ & reminder genital preference not transphobic\dots \\
\bottomrule
\end{tabular}
\caption{Worked input$\to$output examples. The model answers ``Yes'' to every
post (argmax accuracy $0.50$ on this balanced sample: suicide recall $8/8$,
non-suicide specificity $0/8$), yet the logit difference separates the
classes (suicide mean LD $7.46$, range $[3.54, 10.48]$; non-suicide mean LD
$3.26$, range $[0.78, 6.58]$). This affirmative bias is why we report
ranking (AUC) rather than thresholded accuracy throughout.}
\label{tab:examples}
\end{table}

The readout is also robust to prompt phrasing. Run under all three templates
(standard/empathetic/clinical), the same suicidal post gives logit differences
$7.60$/$5.68$/$6.30$ and the same non-suicidal post $0.78$/$2.06$/$1.82$: framing perturbs
the absolute logit difference modestly but never changes the ranking. The affirmative bias
is a property of the model, not of one prompt.

\section{Additional result figures and tables}
\label{app:addfigs}
These figures and tables support claims whose headline numbers are reported in the main
text.

\begin{figure*}[tb]\centering
  \figorbox{figures/fig3_steering.png}
  \caption{Steering raises the readout, but not selectively. Dose--response
  of the Yes$-$No logit difference as the suicidality direction is added to
  non-suicidal posts (layers 12--20). At $c=1$ the injection pushes $100\%$ of posts
  past the class-separating threshold (baseline $8\%$; random direction $14\%$), then
  saturates.}
  \label{fig:steer}
\end{figure*}

\begin{figure*}[tb]\centering
  \figorbox{figures/fig7_validation_scorecard.png}
  \caption{Validation scorecard. Each rung of the causal ladder, scored as the
  suicidality feature (red) against a matched control (grey) on a normalized $0$--$1$ axis
  ($0$ = control/chance floor, $1$ = ideal). A rung holds only when the feature pulls clear
  of its control. Five do; steering does not.}
  \label{fig:scorecard}
\end{figure*}

\begin{table*}[tb]\centering\small
\setlength{\tabcolsep}{5pt}
\begin{tabular}{@{}p{2.5cm}p{3.5cm}p{5.4cm}@{}}
\toprule
Dataset & Form & Role in this study \\
\midrule
DeepSuiMind & Synthetic; three implicit-intent categories
  & Implicit-intent task; implicit-ideation text for the severity analysis \\
Suicide-Watch-232K & Reddit, lemmatized; binary, post-level
  & Primary feature analysis \\
SDCNL & Reddit, raw; binary, post-level
  & Generalization \\
SWMH & Reddit; SuicideWatch vs.\ other mental-health subreddits
  & Generalization; clinical-specificity (separable depression class) \\
UMD (v2) & Reddit, access-controlled; $245$ users, expert ordinal risk (a--d)
  & Validate severity axis vs.\ clinician labels \\
\bottomrule
\end{tabular}
\caption{The five Reddit corpora and their roles in the study.}
\label{tab:datasets}
\end{table*}

\begin{table*}[tb]\centering
\begin{tabular}{llcccl}
\toprule
family & model & params & probe AUC & behav.\ AUC & ablation \\
\midrule
Llama-3.1 & 8B-Instruct$^\dagger$ & 8B & $0.969$ & $0.961$ & $0.98\to0.83$ \\
\midrule
Llama-3.2 & 1B-Instruct & 1B & $0.956$ & $0.488$ & -- (chance) \\
Llama-3.2 & 3B-Instruct & 3B & $0.961$ & $0.900$ & $0.90\to0.48$ \\
\midrule
Qwen2.5 & 0.5B-Instruct & 0.5B & $0.936$ & $0.476$ & -- (chance) \\
Qwen2.5 & 1.5B-Instruct & 1.5B & $0.959$ & $0.894$ & $0.88\to0.45$ \\
Qwen2.5 & 3B-Instruct & 3B & $0.958$ & $0.945$ & $0.94\to0.89$ \\
Qwen2.5 & 7B-Instruct & 7B & $0.975$ & $0.965$ & $0.96\to0.69$ \\
\midrule
Mistral & 7B-Instruct-v0.3 & 7B & $0.976$ & $0.960$ & $0.96\to0.59$ \\
\bottomrule
\end{tabular}
\caption{Cross-model and cross-scale replication ($n=150$/class; probe AUC at the
best layer; ablation = per-layer difference-of-means
direction vs.\ an equal-norm random control). The suicidality representation
(probe) is strong at every scale and family; the behavioral readout emerges with
scale (chance at $\approx$1B in both families, present from $\approx$1.5B), while single-direction
ablation strengthens only gradually (weak at 3B, e.g.\ Qwen2.5-3B $0.94\to0.89$, and
large by 7--8B). $^\dagger$primary model.}
\label{tab:crossmodel}
\end{table*}

\section{Complete evidence table}
\label{sec:evidence}
Table~\ref{tab:summary-full} is the complete form of the main-text scorecard: every
headline result with its matched control and confidence interval, including rows the
main-text table omits (downstream detection, cross-model and cross-scale replication by
family, and the head-level analysis).

\begin{table*}[t]\centering
\setlength{\tabcolsep}{5pt}\footnotesize\renewcommand{\arraystretch}{0.95}
\begin{tabular}{p{0.17\linewidth} p{0.29\linewidth} p{0.25\linewidth} p{0.18\linewidth}}
\toprule
Property & Test & Result & Control / baseline \\
\midrule
Intent NOT represented & behavioral / probe AUC (DeepSuiMind) & $0.557$ / $0.555$--$0.570$ & TF-IDF $0.599$; pos.\ ctrl $1.000$ \\
Decodable & probe AUC, layer 15 & $0.969$ & TF-IDF $0.894$; $p{<}10^{-3}$ \\
Semantic & probe recall, keywords masked & $1.00\to0.74$ & TF-IDF $0.98\to0.29$ \\
Specific & suicide vs.\ depression AUC & $0.744$ $[0.672,0.809]$ & chance $0.50$ \\
Specific (matched neg., ladder) & necessity, suicide-vs-depression & $0.681\to0.466$ (probe $0.796$) & random dir $0.686$; lexical $0.715$ \\
Necessary & per-layer ablation AUC & $0.975\to0.832$ & random dir $0.975$ (drop $0.000$) \\
Low-dimensional & rank-$k$ subspace ablation AUC & $0.963\to0.53$--$0.77$ over $k=1$--$16$ & random rank-$k$ $0.950$--$0.967$ \\
Rank identified & smallest $k$ not above chance & $k^\ast=3,3,10$ across three seeds & sign-blind AUC threshold \\
Sufficiency (not established) & steering flip rate, $c{=}1$ & $100\%$ ($n{=}100$) & random $<1.5$ logits; off-target shift $16.5$ vs.\ on-target $9.6$ \\
General (decode) & transfer AUC, SDCNL / SWMH & $0.963$ / $0.917$ & n/a \\
General (causal) & ablate-232K AUC, SDCNL / SWMH & $0.947\to0.632$ / $0.932\to0.631$ & random dir $\approx0$ \\
Downstream & ablate dir / rank-4, detection F1 & $0.918\to0.690$ / $\to0.000$ & random dir $0.918$ \\
Cross-model (decode) & probe AUC, Qwen / Mistral-Inst & $0.978$ / $0.979$ & TF-IDF/chance \\
Cross-scale & probe AUC, $0.5$--$8$B & $0.94$--$0.98$ & behav.\ from $\approx$1.5B \\
Framework (2nd concept) & necessity, toxicity & $0.910\to0.567$ (probe $0.925$) & random dir $0.913$; lexical $0.802$ \\
Severity axis & projection, neut/dep/impl/expl & $-1.89<0.69<0.95<1.44$ & monotonic \\
Severity (expert) & proj.\ vs.\ UMD risk a/b/c/d & $\rho{=}0.56$; hi-lo AUC $0.81$ & $\rho{=}0.57$ given length; $n{=}245$ users \\
Heads & top single-head knockout & $0.013$, CI $[-0.019,0.051]$ ($n{=}60$) & distributed; $0/1024$ after FDR \\
\bottomrule
\end{tabular}
\caption{Evidence at a glance (complete). CIs are given where a margin is close; every
causal effect is paired with a matched random control. Margins over the lexical baseline
are tested AUC-to-AUC (DeLong + paired bootstrap); the severity readouts by TOST
equivalence; reported $p$ are after a Benjamini--Hochberg correction across the seven
positive rungs (all $q<0.05$).}
\label{tab:summary-full}
\end{table*}

\bibliographystyle{naturemag}
\bibliography{references}

\end{document}